\documentclass{tlp} 
\usepackage{aopmath}
\usepackage{amssymb}

\newtheorem{theo}{Theorem}[section]
\newtheorem{kor}[theo]{Corollary}
\newtheorem{defn}[theo]{Definition}

\newtheorem{bsp}[theo]{Example}

\newcommand{\ground}{\mathsf{ground}}
\newcommand{\body}{\texttt{body}}
\newcommand{\bodyone}{\texttt{body1}}

\newcommand{\domain}{\mathsf{dom}}
\newcommand{\lfp}{\mathsf{lfp}}
\newcommand{\GL}{\mathrm{GL}}
\newcommand{\upa}{\!\uparrow\!}
\newcommand{\Nat}{\mathbb{N}}

\newcommand{\mylabel}[1]{\label{#1}\marginpar{\tiny #1}}

\renewcommand{\mylabel}[1]{\label{#1}}

\begin{document}
\bibliographystyle{acmtrans}

\title[Uniform LP semantics]{A uniform approach to logic programming semantics}


\author[P. Hitzler and M. Wendt]{Pascal Hitzler and Matthias Wendt\\
 Knowledge Representation and Reasoning Group, 
 Artificial Intelligence Institute \\
 Department of Computer Science, 
 Dresden University of Technology \\
 Dresden, Germany\\
 \email{\{phitzler,mw177754\}@inf.tu-dresden.de} \\
} 

\jdate{October 2003}
\pubyear{2003}
\pagerange{\pageref{firstpage}--\pageref{lastpage}}

\maketitle

\begin{abstract}
Part of the theory of logic programming and nonmonotonic reasoning concerns
the study of fixed-point semantics for these paradigms. Several different
semantics have been proposed during the last two decades, and some have been
more successful and acknowledged than others. The rationales behind those
various semantics have been manifold, depending on one's point of view,
which may be that of a programmer or inspired by commonsense reasoning, and
consequently the constructions which lead to these semantics are technically
very diverse, and the exact relationships between them have not yet been
fully understood. In this paper, we present a conceptually new method, based
on level mappings, which allows to provide uniform characterizations of
different semantics for logic programs. We will display our approach by
giving new and uniform characterizations of some of the major semantics,
more particular of the least model semantics for definite programs, of the
Fitting semantics, and of the well-founded semantics. A novel
characterization of the weakly perfect model semantics will also be
provided.
\end{abstract}


\tableofcontents

\section{Introduction}

Negation in logic programming differs from the negation of classical
logic. Indeed, the quest for a satisfactory understanding of negation in
logic programming is still inconclusive --- although the issue has cooled
down a bit recently --- and has proved to be very stimulating for research
activities in computational logic, and in particular amongst knowledge
representation and reasoning researchers concerned with commonsense and
nonmonotonic reasoning. During the last two decades, different
interpretations of negation in logic programming have lead to the
development of a variety of \emph{declarative semantics}, as they are
called. Some early research efforts for establishing a satisfactory
declarative semantics for negation as failure and its variants, as featured
by the resolution-based Prolog family of logic programming systems, have
later on been merged with nonmonotonic frameworks for commonsense reasoning,
culminating recently in the development of so-called answer set programming
systems, like \textsc{smodels} or \textsc{dlv}
\cite{ELMPS97,MT99,Lif02,SNS02}.

Systematically, one can understand Fitting's proposal \cite{Fit85} of a
\emph{Kripke-Kleene semantics} --- also known as \emph{Fitting semantics}
--- as a cornerstone which plays a fundamental r\^ole both for
resolution-based and nonmonotonic reasoning inspired logic
programming. Indeed, his proposal, which is based on a monotonic semantic
operator in Kleene's strong three-valued logic, has been pursued in both
communities, for example by Kunen \cite{Kun87} for giving a semantics for
pure Prolog, and by Apt and Pedreschi \cite{AP93} in their fundamental paper
on termination analysis of negation as failure, leading to the notion of
\emph{acceptable program}. On the other hand, however, Fitting himself
\cite{Fit91,FitTCS}, using a bilattice-based approach which was further
developed by Denecker, Marek, and Truszczynski \cite{DMT00}, tied his
semantics closely to the major semantics inspired by nonmonotonic reasoning,
namely the \emph{stable model semantics} due to Gelfond and Lifschitz
\cite{GL88}, which is based on a nonmonotonic semantic operator, and the
\emph{well-founded semantics} due to van Gelder, Ross, and Schlipf
\cite{GRS91}, originally defined using a different monotonic operator in
three-valued logic together with a notion of unfoundedness.

Another fundamental idea which was recognised in both communities was that
of \emph{stratification}, with the underlying idea of restricting attention
to certain kinds of programs in which recursion through negation is
prevented. Apt, Blair, and Walker \cite{ABW88} proposed a variant of
resolution suitable for these programs, while Przymusinski \cite{Prz88} and
van Gelder \cite{Gel88} generalized the notion to \emph{local}
stratification. Przymusinski \cite{Prz88} developed the \emph{perfect model
semantics} for locally stratified programs, and together with Przymusinska
\cite{PP90} generalized it later to a three-valued setting as the
\emph{weakly perfect model semantics}.

The semantics mentioned so far are defined and characterized using a variety
of different techniques and constructions, including monotonic and
nonmonotonic semantic operators in two- and three-valued logics, program
transformations, level mappings, restrictions to suitable subprograms,
detection of cyclic dependencies etc. Relationships between the semantics
have been established, but even a simple comparison of the respective models
in restricted cases could be rather tedious. So, in this paper, we propose a
methodology which allows to obtain uniform characterizations of all
semantics previously mentioned, and we believe that it will scale up well to
most semantics based on monotonic operators, and also to some nonmonotonic
operators, and to extensions of the logic programming paradigm including
disjunctive conclusions and uncertainty. The characterizations will allow
immediate comparison between the semantics, and once obtained we will easily
be able to make some new and interesting observations, including the fact
that the well-founded semantics can formally be understood as a Fitting
semantics augmented with a form of stratification. Indeed we will note that
from this novel perspective the well-founded semantics captures the idea of
stratification much better than the weakly perfect model semantics, thus
providing a formal explanation for the historic fact that the latter has not
received as much attention as the former.

The main tool which will be employed for our characterizations is the notion
of \emph{level mapping}. Level mappings are mappings from Herbrand bases to
ordinals, i.e. they induce orderings on the set of all ground atoms while
disallowing infinite descending chains. They have been a technical tool in a
variety of contexts, including termination analysis for resolution-based
logic programming as studied by Bezem \cite{Bez89}, Apt and Pedreschi
\cite{AP93}, Marchiori \cite{Mar96}, Pedreschi, Ruggieri, and Smaus
\cite{PRS02}, and others, where they appear naturally since ordinals are
well-orderings. They have been used for defining classes of programs with
desirable semantic properties, e.g. by Apt, Blair, and Walker \cite{ABW88},
Przymusinski \cite{Prz88} and Cavedon \cite{Cav91}, and they are intertwined
with topological investigations of fixed-point semantics in logic
programming, as studied e.g. by Fitting \cite{Fit94,FitTCS}, and by Hitzler
and Seda \cite{Sed95,Sed97,thesis,HS03tcs}. Level mappings are also relevant
to some aspects of the study of relationships between logic programming and
artificial neural networks, as studied by H\"olldobler, Kalinke, and St\"orr
\cite{HKS99} and by Hitzler and Seda \cite{HS00d,HS03}. In our novel
approach to uniform characterizations of different semantics, we will use
them as a technical tool for capturing dependencies between atoms in a
program.

The paper is structured as follows. Section \ref{sec:prelim} contains
preliminaries which are needed to make the paper relatively
self-contained. The subsequent sections contain the announced uniform
characterizations of the least model semantics for definite programs and the
stable model semantics in Section \ref{sec:leaststable}, of the Fitting
semantics in Section \ref{sec:fitting}, of the well-founded semantics in
Section \ref{sec:wf}, and of the weakly perfect model semantics in Section
\ref{sec:wstrat}. Related work will be discussed in Section
\ref{sec:related}, and we close with conclusions and a discussion of further
work in Section \ref{sec:conc}.

Part of this paper was presented at the 25th German Conference on Artificial
Intelligence, KI2002, Aachen, Germany, September 2002 \cite{HW02}.

\bigskip

\emph{Acknowledgement.} We thank Tony Seda for pointing out some flaws in a
previous version of the proof of Theorem \ref{theo:wfchar}.

\section{Preliminaries and Notation}\mylabel{sec:prelim}

A (\emph{normal}) \emph{logic program} is a finite set of (universally
quantified) \emph{clauses} of the form $\forall(A\gets A_1\wedge\dots\wedge
A_n\wedge\lnot B_1\wedge\dots\wedge\lnot B_m)$, commonly written as $A\gets
A_1,\dots,A_n,\lnot B_1,\dots,\lnot B_m$, where $A$, $A_i$, and $B_j$, for
$i=1,\dots,n$ and $j=1,\dots,m$, are atoms over some given first order
language. $A$ is called the \emph{head} of the clause, while the remaining
atoms make up the \emph{body} of the clause, and depending on context, a
body of a clause will be a set of literals (i.e. atoms or negated atoms) or
the conjunction of these literals. Care will be taken that this
identification does not cause confusion. We allow a body, i.e. a
conjunction, to be empty, in which case it always evaluates to true. A
clause with empty body is called a \emph{unit clause} or a \emph{fact}. A
clause is called \emph{definite}, if it contains no negation symbol. A
program is called \emph{definite} if it consists only of definite
clauses. We will usually denote atoms with $A$ or $B$, and literals, which
may be atoms or negated atoms, by $L$ or $K$.

Given a logic program $P$, we can extract from it the components of a first
order language. The corresponding set of ground atoms, i.e. the
\emph{Herbrand base} of the program, will be denoted by $B_P$. For a subset
$I\subseteq B_P$, we set $\lnot I=\{\lnot A\mid A\in I\}$. The set of all
ground instances of $P$ with respect to $B_P$ will be denoted by
$\ground(P)$. For $I\subseteq B_P\cup\lnot B_P$ we say that $A$ is
\emph{true with respect to} (or \emph{in}) $I$ if $A\in I$, we say that $A$
is \emph{false with respect to} (or \emph{in}) $I$ if $\lnot A\in I$, and if
neither is the case, we say that $A$ is \emph{undefined with respect to} (or
\emph{in}) $I$. A (\emph{three-valued} or \emph{partial})
\emph{interpretation} $I$ for $P$ is a subset of $B_P\cup\lnot B_P$ which is
\emph{consistent}, i.e. whenever $A\in I$ then $\lnot A\not\in I$. A body,
i.e. a conjunction of literals, is true in an interpretation $I$ if every
literal in the body is true in $I$, it is false in $I$ if one of its
literals is false in $I$, and otherwise it is undefined in $I$. For a
negative literal $L=\lnot A$ we will find it convenient to write $\lnot L\in
I$ if $A\in I$ and say that $L$ is false in $I$ etc. in this case. By $I_P$
we denote the set of all (three-valued) interpretations of $P$. It is a
complete partial order (cpo) via set-inclusion, i.e. it contains the empty
set as least element, and every ascending chain has a supremum, namely its
union. A \emph{model} of $P$ is an interpretation $I\in I_P$ such that for
each clause $A\gets\body$ we have that $\body$ true in $I$ implies $A$ true
in $I$, and $\body$ undefined in $I$ implies $A$ true or undefined in $I$.
A \emph{total} interpretation is an interpretation $I$ such that no $A\in
B_P$ is undefined in $I$.

For an interpretation $I$ and a program $P$, an \emph{$I$-partial level
mapping} for $P$ is a partial mapping $l:B_P\to\alpha$ with domain
$\domain(l)=\{A\mid A\in I\text{ or }\lnot A\in I\}$, where $\alpha$ is some
(countable) ordinal. We extend every level mapping to literals by setting
$l(\lnot A)=l(A)$ for all $A\in\domain(l)$. A (\emph{total}) \emph{level
mapping} is a total mapping $l:B_P\to\alpha$ for some (countable) ordinal
$\alpha$.

Given a normal logic program $P$ and some $I\subseteq B_P\cup\lnot B_P$, we say
that $U\subseteq B_P$ is an \emph{unfounded set} (\emph{of $P$}) \emph{with
respect to $I$} if each atom $A\in U$ satisfies the following condition: For
each clause $A\gets\body$ in $\ground(P)$ (at least) one of the following
holds.
\begin{description}
\item[(Ui)] Some (positive or negative) literal in $\body$ is false in $I$.
\item[(Uii)] Some (non-negated) atom in $\body$ occurs in $U$.
\end{description}

Given a normal logic program $P$, we define the following operators on
$B_P\cup\lnot B_P$. $T_P(I)$ is the set of all $A\in B_P$ such that there
exists a clause $A\gets\body$ in $\ground(P)$ such that $\body$ is true in
$I$. $F_P(I)$ is the set of all $A\in B_P$ such that for all clauses
$A\gets\body$ in $\ground(P)$ we have that $\body$ is false in $I$. Both
$T_P$ and $F_P$ map elements of $I_P$ to elements of $I_P$. Now define the
operator $\Phi_P: I_P\to I_P$ by
$$
\Phi_P(I) = T_P(I)\cup\lnot F_P(I).
$$ This operator is due to Fitting \cite{Fit85} and is monotonic on the cpo
$I_P$, hence has a least fixed point by the Tarski fixed-point theorem, and
we can obtain this fixed point by defining, for each monotonic operator $F$,
that $F\upa 0=\emptyset$, $F\upa(\alpha+1)=F(F\upa\alpha)$ for any ordinal
$\alpha$, and $F\upa\beta=\bigcup_{\gamma<\beta}F\upa\gamma$ for any limit
ordinal $\beta$, and the least fixed point $\lfp(F)$ of $F$ is obtained as
$F\upa\alpha$ for some ordinal $\alpha$. The least fixed point of $\Phi_P$
is called the \emph{Kripke-Kleene model} or \emph{Fitting model} of $P$,
determining the \emph{Fitting semantics} of $P$.

\begin{bsp}\mylabel{bsp:fit}
Let $P$ be the program consisting of the two clauses $p\gets p$ and
$q\gets\lnot r$. Then $\Phi_P\upa 1=\{\lnot r\}$, and $\Phi_P\upa
2=\{q,\lnot r\}=\Phi_P\upa 3$ is the Fitting model of $P$.
\end{bsp}

Now, for $I\subseteq B_P\cup\lnot B_P$, let $U_P(I)$ be the greatest
unfounded set (of $P$) with respect to $I$, which always exists due to van
Gelder, Ross, and Schlipf \cite{GRS91}. Finally, define
$$
W_P(I) = T_P(I)\cup\lnot U_P(I)
$$ for all $I\subseteq B_P\cup\lnot B_P$. The operator $W_P$, which operates
on the cpo $B_P\cup\lnot B_P$, is due to van Gelder et al. \cite{GRS91} and
is monotonic, hence has a least fixed point by the Tarski fixed-point
theorem, as above for $\Phi_P$. It turns out that $W_P\upa\alpha$ is in
$I_P$ for each ordinal $\alpha$, and so the least fixed point of $W_P$ is
also in $I_P$ and is called the \emph{well-founded model} of $P$, giving the
\emph{well-founded semantics} of $P$.

\begin{bsp}\mylabel{bsp:wf}
Let $P$ be the program consisting of the following clauses.
\begin{center}
\begin{minipage}[t]{4cm}
\begin{displaymath}
\begin{array}{lcl}
s & \gets & q \\
q & \gets & \lnot p \\
p & \gets & p \\
r & \gets & \lnot r
\end{array}
\end{displaymath}
\end{minipage}
\end{center}
Then $\{p\}$ is the largest unfounded set of $P$ with respect to $\emptyset$
and we obtain 
\begin{center}
\begin{minipage}[t]{5.55cm} 
\begin{displaymath}
\begin{array}{lcl}
W_P\upa 1 &=& \{\lnot p\},\\
W_P\upa 2 &=& \{\lnot p,q\}, \qquad\text{and}\\
W_P\upa 3 &=& \{\lnot p,q,s\} \\&=&W_P\upa 4.
\end{array}
\end{displaymath}
\end{minipage}
\end{center}

\end{bsp}

Given a program $P$, we define the operator $T_P^\plus$ on subsets of $B_P$
by $T_P^\plus(I)=T_P(I\cup\lnot (B_P\setminus I))$.  It is well-known that
for definite programs this operator is monotonic on the set of all subsets
of $B_P$, with respect to subset inclusion. Indeed it is Scott-continuous
\cite{Llo88,AJ94,SLG94} and, via Kleene's fixed-point theorem, achieves its
least fixed point $M$ as the supremum of the iterates $T_P^\plus\upa n$ for
$n\in\Nat$. So $M=\lfp(T_P^\plus)=T_P^\plus\upa\omega$ is \emph{the least
two-valued model} of $P$. In turn, we can identify $M$ with the total
interpretation $M\cup\lnot (B_P\setminus M)$, which we will call the
\emph{definite} (\emph{partial}) \emph{model} of $P$.

\begin{bsp}\mylabel{bsp:definite}
Let $P$ be the program consisting of the clauses

\begin{center}
\begin{minipage}[t]{5cm}
\begin{displaymath}
\begin{array}{lcl}
p(0) &\gets\\
p(s(X)) &\gets & p(X),
\end{array}
\end{displaymath}
\end{minipage}
\end{center}
where $X$ denotes a variable and $0$ a constant symbol. Write $s^n(0)$ for
the term $s(\cdots s(0)\cdots)$ in which the symbol $s$ appears $n$
times. Then
$$T_P^\plus\upa n=\left\{p\left(s^k(0)\right)\mid k<n\right\}$$
for all $n\in\Nat$ and $\{p(s^n(0))\mid n\in\Nat\}$ is the least two-valued
model of $P$.
\end{bsp}

In order to avoid confusion, we will use the following terminology: the
notion of \emph{interpretation} will by default denote consistent subsets of
$B_P\cup\lnot B_P$, i.e. interpretations in three-valued logic. We will
sometimes emphasize this point by using the notion \emph{partial
interpretation}. By \emph{two-valued interpretations} we mean subsets of
$B_P$. Given a partial interpretation $I$, we set $I^\plus=I\cap B_P$ and
$I^\minus= \{A\in B_P\mid \lnot A\in I\}$. Each two-valued interpretation $I$
can be identified with the partial interpretation $I'=I\cup\lnot (B_P\setminus
I)$. Both, interpretations and two-valued interpretations, are ordered by
subset inclusion. We note however, that these two orderings differ: If
$I\subseteq B_P$, for example, then $I'$ is always a maximal element in the
ordering for partial interpretations, while $I$ is in general not maximal as
a two-valued interpretation. The two orderings correspond to the knowledge-
and the truth-ordering due to Fitting \cite{Fit91}.

There is a semantics using two-valued logic, the stable model semantics due
to Gelfond and Lifschitz \cite{GL88}, which is intimately related to the
well-founded semantics. Let $P$ be a normal program, and let $M\subseteq
B_P$ be a set of atoms. Then we define $P/M$ to be the (ground) program
consisting of all clauses $A\gets A_1,\dots,A_n$ for which there is a clause
$A\gets A_1,\dots,A_n,\lnot B_1,\dots,\lnot B_m$ in $\ground(P)$ with
$B_1,\dots,B_m\not\in M$. Since $P/M$ does no longer contain negation, it
has a least two-valued model $T_{P/M}^\plus\upa\omega$. For any two-valued
interpretation $I$ we can therefore define the operator
$\GL_P(I)=T_{P/I}^\plus\upa\omega$, and call $M$ a \emph{stable model} of the
normal program $P$ if it is a fixed point of the operator $\GL_P$, i.e. if
$M=\GL_P(M)=T_{P/M}^\plus\upa\omega$. As it turns out, the operator $\GL_P$ is
in general not monotonic for normal programs $P$. However it is
\emph{antitonic}, i.e. whenever $I\subseteq J\subseteq B_P$ then
$\GL_P(J)\subseteq\GL_P(I)$. As a consequence, the operator $\GL_P^2$,
obtained by applying $\GL_P$ twice, is monotonic and hence has a least fixed
point $L_P$ and a greatest fixed point $G_P$. Van Gelder \cite{Gel89} has
shown that $\GL_P(L_P)=G_P$, $L_P=\GL_P(G_P)$, and that $L_P\cup\lnot
(B_P\setminus G_P)$ coincides with the well-founded model of $P$. This is
called the \emph{alternating fixed point characterization} of the
well-founded semantics.

\begin{bsp}\mylabel{bsp:afp}
Consider the program $P$ from Example \ref{bsp:wf}. The subprogram $Q$
consisting of the first three clauses of the program $P$ has stable model
$M=\{s,q\}$, which can be verified by noting that $Q/M$ consists of the
clauses

\begin{center}
\begin{minipage}[t]{3.8cm}
\begin{displaymath}
\begin{array}{lcl}
s &\gets & q\\
q &\gets\\
p &\gets& p,
\end{array}
\end{displaymath}
\end{minipage}
\end{center}
and has $M$ as its least two-valued model.

For the program $P$ we obtain

\begin{center}
\begin{minipage}[t]{7.7cm}
\begin{displaymath}
\begin{array}{lcl}
\GL_P(\emptyset) &=& \{q,s,r\},\\
\GL_P(\{q,s,r\}) &=& \{q,s\} \\ &=&\GL_P^2(\{q,s\}),\qquad\text{and}\\
\GL_P(B_P) &=& \emptyset.
\end{array}
\end{displaymath}
\end{minipage}
\end{center}

So $L_P=\{q,s\}$ while $G_P=\{q,s,r\}$, and $L_P\cup\lnot (B_P\setminus
G_P)=\{q,s,\lnot p\}$ is the well-founded model of $P$. 
\end{bsp}

\section{Least and Stable Model Semantics}\mylabel{sec:leaststable}

The most fundamental semantics in logic programming is based on the fact
mentioned above that the operator $T_P^\plus$ has a least fixed point
$M=T_P^\plus\upa\omega$ whenever $P$ is definite. The two-valued interpretation
$M$ turns out to be the least two-valued model of the program, and is
therefore canonically the model which should be considered for definite
programs. Our first result characterizes the least model using level
mappings, and serves to convey the main ideas underlying our method. It is a
straightforward result but has, to the best of our knowledge, not been noted
before.

\begin{theo}\mylabel{theo:defleast}
Let $P$ be a definite program. Then there is a unique two-valued model $M$ of
$P$ for which there exists a (total) level mapping $l: B_P\to\alpha$ such that
for each atom $A\in M$ there exists a clause $A\gets A_1,\dots,A_n$ in
$\ground(P)$ with $A_i\in M$ and $l(A)>l(A_i)$ for all
$i=1,\dots,n$. Furthermore, $M$ is the least two-valued model of $P$.
\end{theo}

\begin{proof}
Let $M$ be the least two-valued model $T_P^\plus\upa\omega$, choose
$\alpha=\omega$, and define $l:B_P\to\alpha$ by setting $l(A)=\min\{n\mid
A\in T_P^\plus\upa (n+1)\}$, if $A\in M$, and by setting $l(A)=0$, if $A\not\in
M$. From the fact that $\emptyset \subseteq T_P^\plus\upa 1 \subseteq \ldots
\subseteq T_P^\plus\upa n \subseteq \ldots \subseteq T_P^\plus\upa\omega=\bigcup_m
T_P^\plus\upa m$, for each $n$, we see that $l$ is well-defined and that the
least model $T_P^\plus\upa\omega$ for $P$ has the desired properties.

Conversely, if $M$ is a two-valued model for $P$ which satisfies the given
condition for some mapping $l:B_P\to\alpha$, then it is easy to show, by
induction on $l(A)$, that $A\in M$ implies $A\in T_P^\plus\upa (l(A)+1)$. This
yields that $M\subseteq T_P^\plus\upa\omega$, and hence that $M=
T_P^\plus\upa\omega$ by minimality of the model $T_P^\plus\upa\omega$.
\end{proof}

\begin{bsp}\mylabel{bsp:definite2}
For the program $P$ from Example \ref{bsp:definite} we obtain $l(p(s^n(0)))=
n$ for the level mapping $l$ defined in the proof of Theorem
\ref{theo:defleast}.
\end{bsp}

The proof of Theorem \ref{theo:defleast} can serve as a blueprint for
obtaining characterizations if the semantics under consideration is based on
the least fixed point of a monotonic operator $F$, and indeed our results
for the Fitting semantics and the well-founded semantics, Theorems
\ref{theo:fitchar} and \ref{theo:wfchar}, together with their proofs, follow
this scheme. In one direction, levels are assigned to atoms $A$ according to
the least ordinal $\alpha$ such that $A$ is not undefined in
$F\upa(\alpha+1)$, and dependencies between atoms of some level and atoms of
lower levels are captured by the nature of the considered operator, which
will certainly vary from case to case. In Theorem \ref{theo:defleast}, the
condition thus obtained suffices for uniquely determining the least model,
whereas in other cases which we will study later, so for the Fitting
semantics and the well-founded semantics, the level mapping conditions will
not suffice for unique characterization of the desired model. However, the
desired model will in each case turn out to be the greatest among all models
satisfying the given conditions. So in these cases it will remain to show,
by transfinite induction on the level of some given atom $A$, that the truth
value assigned to $A$ by any model satisfying the given conditions is also
assigned to $A$ by $F\upa(l(A)+1)$, which at the same time proves that
$\lfp(F)$ is the greatest model satisfying the given conditions. For the
proof of Theorem \ref{theo:defleast}, the proof method just described can be
applied straightforwardly, however for more sophisticated operators may
become technically challenging on the detailed level.

We now turn to the stable model semantics, which in the case of programs
with negation has come to be the major semantics based on two-valued
logic. The following characterization is in the spirit of our proposal, and
is due to Fages \cite{Fag94}. It is striking in its similarity to the
characterization of the least model for definite programs in Theorem
\ref{theo:defleast}. For completeness of our exhibition, we include a proof
of the fact.

\begin{theo}\mylabel{theo:stablechar}
Let $P$ be normal. Then a two-valued model $M\subseteq B_P$ of $P$ is a
stable model of $P$ if and only if there exists a (total) level mapping
$l:B_P\to\alpha$ such that for each $A\in M$ there exists $A\gets
A_1,\dots,A_n,\lnot B_1,\dots,\lnot B_m$ in $\ground(P)$ with $A_i\in M$,
$B_j\not\in M$, and $l(A)>l(A_i)$ for all $i=1,\dots,n$ and $j=1,\dots,m$.
\end{theo}

\begin{proof}
Let $M$ be a stable model of $P$,
i.e. $\GL_P(M)=T_{P/M}^\plus\upa\omega=M$. Then $M$ is the least model for
$P/M$, hence is also a model for $P$, and, by Theorem \ref{theo:defleast},
satisfies the required condition with respect to any level mapping $l$ with
$l(A)=\min\{n\mid A\in T_{P/M}\upa (n+1)\}$ for each $A\in M$. Conversely,
let $M$ be a model which satisfies the condition in the statement of the
theorem. Then, for every $A\in M$, there is a clause $C$ in $\ground(P)$ of
the form $A\gets A_1,\dots,A_n,\lnot B_1,\dots,\lnot B_k$ such that the body
of $C$ is true in $M$ and satisfies $l(A)>l(A_i)$ for all $i=1,\dots,n$. But
then, for every $A\in M$, there is a clause $A\gets A_1,\dots,A_n$ in $P/M$
whose body is true in $M$ and such that $l(A)>l(A_i)$ for all
$i=1,\dots,n$. By Theorem \ref{theo:defleast}, this means that $M$ is the
least model for $P/M$, that is, $M=T_{P/M}^\plus\upa\omega=\GL(M)$.
\end{proof}

The proof of Theorem \ref{theo:stablechar} just given partly follows the
proof scheme discussed previously, by considering the monotonic operator
$T_{P/M}^\plus$, which is used for defining stable models.

\begin{bsp}\mylabel{bsp:stable}
Recall the program $P$ from Example \ref{bsp:wf}, and consider the program
$Q$ consisting of the first three clauses of $P$. We already noted in
Example \ref{bsp:afp} that $Q$ has stable model $\{s,q\}$. A corresponding
level mapping, as defined in the proof of Theorem \ref{theo:stablechar},
satisfies $l(q)=0$ and $l(s)=1$, while $l(p)$ can be an arbitrary value.
\end{bsp}

\section{Fitting Semantics}\mylabel{sec:fitting}

We next turn to the Fitting semantics. Following the proof scheme which we
described in Section \ref{sec:leaststable}, we expect levels $l(A)$ to be
assigned to atoms $A$ such that $l(A)$ is the least $\alpha$ such that $A$
is not undefined in $\Phi_P\upa(\alpha+1)$. An analysis of the operator
$\Phi_P$ eventually yields the following conditions.

\begin{defn}\mylabel{def:fit}
Let $P$ be a normal logic program, $I$ be a model of $P$, and $l$ be an
$I$-partial level mapping for $P$. We say that $P$ \emph{satisfies} (F)
\emph{with respect to $I$ and $l$}, if each $A\in\domain(l)$ satisfies one of
the following conditions.
\begin{enumerate}
\item[(Fi)] $A\in I$ and there exists a clause $A\gets L_1,\dots, L_n$ in
$\ground(P)$ with $L_i\in I$ and $l(A)>l(L_i)$ for all
$i$.
\item[(Fii)] $\lnot A\in I$ and for each clause $A\gets L_1,\dots, L_n$ in
$\ground(P)$ there exists $i$ with $\lnot L_i\in I$ and $l(A)>l(L_i)$.
\end{enumerate}
If $A\in\domain(l)$ satisfies (Fi), then we say that $A$ \emph{satisfies} (Fi)
\emph{with respect to $I$ and $l$}, and similarly if $A\in\domain(l)$ satisfies
(Fii).
\end{defn}

We note that condition (Fi) is stronger than the condition used for
characterizing stable models in Theorem \ref{theo:stablechar}. The proof of
the next theorem closely follows our proof scheme.

\begin{theo}\mylabel{theo:fitchar}
Let $P$ be a normal logic program with Fitting model $M$. Then $M$ is the
greatest model among all models $I$, for which there exists an $I$-partial
level mapping $l$ for $P$ such that $P$ satisfies (F) with respect to $I$
and $l$.
\end{theo}

\begin{proof}
Let $M_P$ be the Fitting model of $P$ and define the $M_P$-partial level
mapping $l_P$ as follows: $l_P(A)=\alpha$, where $\alpha$ is the least
ordinal such that $A$ is not undefined in $\Phi_P\upa(\alpha+1)$. The proof
will be established by showing the following facts: (1) $P$ satisfies (F)
with respect to $M_P$ and $l_P$. (2) If $I$ is a model of $P$ and $l$ an
$I$-partial level mapping such that $P$ satisfies (F) with respect to $I$
and $l$, then $I\subseteq M_P$.

(1) Let $A\in\domain(l_P)$ and $l_P(A)=\alpha$. We consider two cases. 

(Case i) If $A\in M_P$, then $A\in T_P(\Phi_P\upa\alpha)$, hence there
exists a clause $A\gets\body$ in $\ground(P)$ such that $\body$ is true in
$\Phi_P\upa\alpha$. Thus, for all $L_i\in\body$ we have that $L_i\in
\Phi_P\upa\alpha$, and hence $l_P(L_i)<\alpha$ and $L_i\in M_P$ for all
$i$. Consequently, $A$ satisfies (Fi) with respect to $M_P$ and $l_P$. 

(Case ii) If $\lnot A\in M_P$, then $A\in F_P(\Phi_P\upa\alpha)$, hence for
all clauses $A\gets\body$ in $\ground(P)$ there exists $L\in\body$ with
$\lnot L\in\Phi_P\upa\alpha$ and $l_P(L)<\alpha$, hence $\lnot L\in
M_P$. Consequently, $A$ satisfies (Fii) with respect to $M_P$ and $l_P$, and
we have established that fact (1) holds.

(2) We show via transfinite induction on $\alpha=l(A)$, that whenever $A\in
I$ (respectively, $\lnot A\in I$), then $A\in\Phi_P\upa(\alpha+1)$
(respectively, $\lnot A\in\Phi_P\upa(\alpha+1)$). For the base case, note
that if $l(A) = 0$, then $A\in I$ implies that $A$ occurs as the head of a
fact in $\ground(P)$, hence $A\in\Phi_P\upa 1$, and $\lnot A\in I$ implies
that there is no clause with head $A$ in $\ground(P)$, hence $\lnot
A\in\Phi_P\upa 1$. So assume now that the induction hypothesis holds for all
$B\in B_P$ with $l(B)<\alpha$. We consider two cases.

(Case i) If $A\in I$, then it satisfies (Fi) with respect to $I$ and
$l$. Hence there is a clause $A\gets\body$ in $\ground(P)$ such that
$\body\subseteq I$ and $l(K)<\alpha$ for all $K\in\body$. Hence
$\body\subseteq M_P$ by induction hypothesis, and since $M_P$ is a model of
$P$ we obtain $A\in M_P$. 

(Case ii) If $\lnot A\in I$, then $A$ satisfies (Fii) with respect to $I$
and $l$. Hence for all clauses $A\gets\body$ in $\ground(P)$ we have that
there is $K\in\body$ with $\lnot K\in I$ and $l(K)<\alpha$. Hence for all
these $K$ we have $\lnot K\in M_P$ by induction hypothesis, and consequently
for all clauses $A\gets\body$ in $\ground(P)$ we obtain that $\body$ is
false in $M_P$. Since $M_P=\Phi_P(M_P)$ is a fixed point of the
$\Phi_P$-operator, we obtain $\lnot A\in M_P$. This establishes fact (2) and
concludes the proof.
\end{proof}

\begin{bsp}\mylabel{bsp:fit2}
Consider the program $P$ from Example \ref{bsp:fit}. Then the level mapping
$l$, as defined in the proof of Theorem \ref{theo:fitchar}, satsifies
$l(r)=0$ and $l(q)=1$.
\end{bsp}

It is interesting to consider the special case where the Fitting model is
total. Programs with this property are called \emph{$\Phi$-accessible}
\cite{HS99a,HS03tcs}, and include e.g. the acceptable programs due to Apt and
Pedreschi \cite{AP93}.

\begin{kor}\mylabel{kor:fittotal}
A normal logic program $P$ has a total Fitting model if and only if there is
a total model $I$ of $P$ and a (total) level mapping $l$ for $P$ such that
$P$ satisfies (F) with respect to $I$ and $l$.
\end{kor}

The result follows immediately as a special case of Theorem
\ref{theo:fitchar}, and is closely related to results reported in
\cite{HS99a,HS03tcs}. The reader familiar with acceptable programs will also
note the close relationship between Corollary \ref{kor:fittotal} and the
defining conditions for acceptable programs. Indeed, the theorem due to Apt
and Pedreschi \cite{AP93}, which says that every acceptable program has a
total Fitting model, follows without any effort from our result. It also
follows immediately, by comparing Corollary \ref{kor:fittotal} and Theorem
\ref{theo:stablechar}, that a total Fitting model is always stable, which is
a well-known fact.

\section{Well-Founded Semantics}\mylabel{sec:wf}

The characterization of the well-founded model again closely follows our
proof scheme. Before discussing this, though, we will take a short detour
which will eventually reveal a surprising fact about the well-founded
semantics: From our new perspective the well-founded semantics can be
understood as a stratified version of the Fitting semantics.

Let us first recall the definition of a (locally) stratified program, due to
Apt, Blair, Walker, and Przymusinski \cite{ABW88,Prz88}: A normal logic
program is called \emph{locally stratified} if there exists a (total) level
mapping $l:B_P\to\alpha$, for some ordinal $\alpha$, such that for each
clause $A\gets A_1,\dots,A_n, \lnot B_1,\dots,\lnot B_m$ in $\ground(P)$ we
have that $l(A)\geq l(A_i)$ and $l(A)>l(B_j)$ for all $i=1,\dots,n$ and
$j=1,\dots,m$.

The notion of (locally) stratifed program, as already mentioned in the
introduction, was developed with the idea of \emph{preventing recursion
through negation}, while allowing recursion through positive dependencies.
There exist locally stratified programs which do not have a total Fitting
model and vice versa. Indeed, the program consisting of the single clause
$p\gets p$ is locally stratified but $p$ remains undefined in the Fitting
model. Conversely, the program consisting of the two clauses $q\gets$ and
$q\gets\lnot q$ is not locally stratified but its Fitting model assigns to
$q$ the truth value \emph{true}. 

By comparing Definition \ref{def:fit} with the definition of locally
stratified programs, we notice that condition (Fii) requires a strict
decrease of level between the head and a literal in the rule, independent of
this literal being positive or negative. But, on the other hand, condition
(Fii) imposes no further restrictions on the remaining body literals, while
the notion of local stratification does. These considerations motivate the
substitution of condition (Fii) by the condition (WFii), as given in the
following definition.

\begin{defn}\mylabel{def:wfchar}
Let $P$ be a normal logic program, $I$ be a model of $P$, and $l$ be an
$I$-partial level mapping for $P$. We say that \emph{$P$ satisfies} (WF)
\emph{with respect to $I$ and $l$}, if each $A\in\domain(l)$ satisfies one of
the following conditions.
\begin{enumerate}
\item[(WFi)] $A\in I$ and there exists a clause $A\gets L_1,\dots, L_n$ in
$\ground(P)$ with $L_i\in I$ and $l(A)>l(L_i)$ for all
$i$.
\item[(WFii)] $\lnot A\in I$ and for each clause $A\gets A_1,\dots, A_n,\lnot
B_1,\dots,\lnot B_m$ in $\ground(P)$ (at least) one of the
following conditions holds:
\begin{enumerate}
\item[(WFiia)] There exists $i\in\{1,\dots,n\}$ with $\lnot A_i\in I$ and
$l(A)\geq l(A_i)$.
\item[(WFiib)] There exists $j\in\{1,\dots,m\}$ with $B_j\in I$ and
  $l(A)>l(B_j)$.  
\end{enumerate}
\end{enumerate}
If $A\in\domain(l)$ satisfies (WFi), then we say that $A$ \emph{satisfies} (WFi)
\emph{with respect to $I$ and $l$}, and similarly if $A\in\domain(l)$ satisfies
(WFii).
\end{defn}

We note that conditions (Fi) and (WFi) are identical. Indeed, replacing (WFi)
by a stratified version such as the following seems not satisfactory.
\begin{enumerate}
\item[(SFi)] $A\in I$ and there exists a clause $A\gets A_1,\dots, A_n,\lnot
B_1,\dots,\lnot B_m$ in $\ground(P)$ with $A_i,B_j\in I$,
$l(A)\geq l(A_i)$, and $l(A)>l(B_j)$ for all $i$ and $j$.
\end{enumerate}
If we replace condition (WFi) by condition (SFi), then it is not guaranteed
that for any given program there is a greatest model satisfying the desired
properties: Consider the program consisting of the two clauses $p\gets p$
and $q\gets\lnot p$, and the two (total) models $\{p,\lnot q\}$ and $\{\lnot
p,q\}$, which are incomparable, and the level mapping $l$ with $l(p)=0$ and
$l(q)=1$. A detailed analysis of condition (SFi) in the context of our
approach can be found in \cite{Hit03ki}.

So, in the light of Theorem \ref{theo:fitchar}, Definition \ref{def:wfchar}
should provide a natural ``stratified version'' of the Fitting
semantics. And indeed it does, and furthermore, the resulting semantics
coincides with the well-founded semantics, which is a very satisfactory
result. The proof of the fact again follows our proof scheme, but is
slightly more involved due to the necessary treatment of unfounded sets. 

\begin{theo}\mylabel{theo:wfchar}
Let $P$ be a normal logic program with well-founded model $M$. Then $M$ is
the greatest model among all models $I$, for which there exists an
$I$-partial level mapping $l$ for $P$ such that $P$ satisfies (WF) with
respect to $I$ and $l$.
\end{theo}

\begin{proof}
Let $M_P$ be the well-founded model of $P$ and define the $M_P$-partial
level mapping $l_P$ as follows: $l_P(A)=\alpha$, where $\alpha$ is the least
ordinal such that $A$ is not undefined in $W_P\upa(\alpha+1)$. The proof
will be established by showing the following facts: (1) $P$ satisfies (WF)
with respect to $M_P$ and $l_P$. (2) If $I$ is a model of $P$ and $l$ an
$I$-partial level mapping such that $P$ satisfies (WF) with respect to $I$
and $l$, then $I\subseteq M_P$.

(1) Let $A\in\domain(l_P)$ and $l_P(A)=\alpha$. We consider two cases. 

(Case i) If $A\in M_P$, then $A\in T_P(W_P\upa\alpha)$, hence there exists a
clause $A\gets\body$ in $\ground(P)$ such that $\body$ is true in
$W_P\upa\alpha$. Thus, for all $L_i\in\body$ we have that $L_i\in
W_P\upa\alpha$. Hence, $l_P(L_i)<\alpha$ and $L_i\in M_P$ for all
$i$. Consequently, $A$ satisfies (WFi) with respect to $M_P$ and $l_P$.

(Case ii) If $\lnot A\in M_P$, then $A\in U_P(W_P\upa\alpha)$, i.e. $A$ is
contained in the greatest unfounded set of $P$ with respect to
$W_P\upa\alpha$. Hence for each clause $A\gets\body$ in $\ground(P)$, at
least one of (Ui) or (Uii) holds for this clause with respect to
$W_P\upa\alpha$ and the unfounded set $U_P(W_P\upa\alpha)$. If (Ui) holds,
then there exists some literal $L\in\body$ with $\lnot L\in W_P\upa\alpha$.
Hence $l_P(L)<\alpha$ and condition (WFiib) holds relative to $M_P$ and
$l_P$ if $L$ is an atom, or condition (WFiia) holds relative to $M_P$ and
$l_P$ if $L$ is a negated atom. On the other hand, if (Uii) holds, then some
(non-negated) atom $B$ in $\body$ occurs in $U_P(W_P\upa\alpha)$. Hence
$l_P(B)\leq l_P(A)$ and $A$ satisfies (WFiia) with respect to $M_P$ and
$l_P$. Thus we have established that fact (1) holds.

(2) We show via transfinite induction on $\alpha=l(A)$, that whenever $A\in
I$ (respectively, $\lnot A\in I$), then $A\in W_P\upa(\alpha+1)$
(respectively, $\lnot A\in W_P\upa(\alpha+1)$). For the base case, note that
if $l(A)=0$, then $A\in I$ implies that $A$ occurs as the head of a fact in
$\ground(P)$. Hence, $A\in W_P\upa 1$. If $\lnot A\in I$, then consider the
set $U$ of all atoms $B$ with $l(B)=0$ and $\lnot B\in I$. We show that $U$
is an unfounded set of $P$ with respect to $W_P\upa 0$, and this suffices
since it implies $\lnot A\in W_P\upa 1$ by the fact that $A\in U$. So let
$C\in U$ and let $C\gets\body$ be a clause in $\ground(P)$. Since $\lnot
C\in I$, and $l(C)=0$, we have that $C$ satisfies (WFiia) with respect to
$I$ and $l$, and so condition (Uii) is satisfied showing that $U$ is an
unfounded set of $P$ with respect to $I$. Assume now that the induction
hypothesis holds for all $B\in B_P$ with $l(B)<\alpha$. We consider two
cases.

(Case i) If $A\in I$, then it satisfies (WFi) with respect to $I$ and
$l$. Hence there is a clause $A\gets\body$ in $\ground(P)$ such that
$\body\subseteq I$ and $l(K)<\alpha$ for all $K\in\body$. Hence
$\body\subseteq W_P\upa\alpha$, and we obtain $A\in T_P(W_P\upa\alpha)$ as
required. 

(Case ii) If $\lnot A\in I$, consider the set $U$ of all atoms $B$ with
$l(B)=\alpha$ and $\lnot B\in I$. We show that $U$ is an unfounded set of
$P$ with respect to $W_P\upa\alpha$, and this suffices since it implies
$\lnot A\in W_P\upa (\alpha+1)$ by the fact that $A\in U$. So let $C\in U$
and let $C\gets\body$ be a clause in $\ground(P)$. Since $\lnot C\in I$, we
have that $C$ satisfies (WFii) with respect to $I$ and $l$. If there is a
literal $L\in\body$ with $\lnot L\in I$ and $l(L)<l(C)$, then by the
induction hypothesis we obtain $\lnot L\in W_P\upa\alpha$, so condition (Ui)
is satisfied for the clause $C\gets\body$ with respect to $W_P\upa\alpha$
and $U$. In the remaining case we have that $C$ satisfies condition (WFiia),
and there exists an atom $B\in\body$ with $\lnot B\in I$ and $l(B)=l(C)$.
Hence, $B\in U$ showing that condition (Uii) is satisfied for the clause
$C\gets\body$ with respect to $W_P\upa\alpha$ and $U$. Hence $U$ is an
unfounded set of $P$ with respect to $W_P\upa\alpha$.
\end{proof}

\begin{bsp}\mylabel{bsp:wf2}
Consider the program $P$ from Example \ref{bsp:wf}. With notation from the
proof of Theorem \ref{theo:wfchar}, we obtain $l(p)=0$, $l(q)=1$, and
$l(s)=2$. 
\end{bsp}

As a special case, we consider programs with total well-founded model. The
following corollary follows without effort from Theorem \ref{theo:wfchar}.

\begin{kor}\mylabel{kor:wftotal}
A normal logic program $P$ has a total well-founded model if and only if
there is a total model $I$ of $P$ and a (total) level mapping $l$ such that
$P$ satisfies (WF) with respect to $I$ and $l$.
\end{kor}

As a further example for the application of our proof scheme, we use Theorem
\ref{theo:wfchar} in order to prove a result by van Gelder \cite{Gel89}
which we mentioned in the introduction, concerning the alternating
fixed-point characterization of the well-founded semantics. Let us first
introduce some temporary notation, where $P$ is an arbitrary program.

\begin{displaymath}
\begin{array}{lcll}
L_0 &= &\emptyset\\
G_0 &= &B_P\\
L_{\alpha+1} &= &\GL_P(G_\alpha)&\text{ for any ordinal $\alpha$}\\
G_{\alpha+1} &= &\GL_P(L_\alpha)&\text{ for any ordinal $\alpha$}\\
L_\alpha &= &\bigcup_{\beta<\alpha} L_\beta&\text{ for limit ordinal
  $\alpha$}\\
G_\alpha &= &\bigcap_{\beta<\alpha} G_\beta&\text{ for limit ordinal
  $\alpha$} 
\end{array}
\end{displaymath}

Since $\emptyset\subseteq B_P$, we obtain $L_0\subseteq L_1\subseteq
G_1\subseteq G_0$ and, by transfinite induction, it can easily be shown that
$L_\alpha\subseteq L_\beta\subseteq G_\beta\subseteq G_\alpha$ whenever
$\alpha\leq\beta$. In order to apply our proof scheme, we need to detect a
monotonic operator, or at least some kind of monotonic construction,
underlying the alternative fixed-point characterization. The assignment
$(L_\alpha,G_\alpha)\mapsto (L_{\alpha+1},G_{\alpha+1})$, using the
temporary notation introduced above, will serve for this purpose. The proof
of the following theorem is based on it and our general proof scheme, with
modifications where necessary, for example for accomodating the fact that
$G_{\alpha+1}$ is not defined using $G_\alpha$, but rather $L_{\alpha}$, and
that we work with the complements $B_P\setminus G_\alpha$ instead of the
sets $G_\alpha$.

\begin{theo}\mylabel{theo:altfp}
Let $P$ be a normal program. Then $M=L_P\cup\lnot (B_P\setminus G_P)$ is the
well-founded model of $P$.
\end{theo}

\begin{proof}
First, we define an $M$-partial level mapping $l$. For convenience, we will
take as image set of $l$, pairs $(\alpha,n)$ of ordinals, where
$n\leq\omega$, with the lexicographic ordering. This can be done without
loss of generality because any set of pairs of ordinals, lexicographically
ordered, is certainly well-ordered and therefore order-isomorphic to an
ordinal. For $A\in L_P$, let $l(A)$ be the pair $(\alpha,n)$, where $\alpha$
is the least ordinal such that $A\in L_{\alpha+1}$, and $n$ is the least
ordinal such that $A\in T_{P/G_\alpha}\upa (n+1)$. For $B\not\in G_P$, let
$l(B)$ be the pair $(\beta,\omega)$, where $\beta$ is the least ordinal such
that $B\not\in G_{\beta+1}$. We show next by transfinite induction that $P$
satisfies (WF) with respect to $M$ and $l$.

Let $A\in L_1=T_{P/B_P}\upa\omega$. Since $P/B_P$ consists of exactly all
clauses from $\ground(P)$ which contain no negation, we have that $A$ is
contained in the least two-valued model for a definite subprogram of $P$,
namely $P/B_P$, and (WFi) is satisfied by Theorem \ref{theo:defleast}. Now
let $\lnot B\in \lnot (B_P\setminus G_P)$ be such that $B\in (B_P\setminus
G_1)=B_P\setminus T_{P/\emptyset}\upa\omega$. Since $P/\emptyset$ contains
all clauses from $\ground(P)$ with all negative literals removed, we obtain
that each clause in $\ground(P)$ with head $B$ must contain a positive body
literal $C\not\in G_1$, which, by definition of $l$, must have the same
level as $B$, hence (WFiia) is satisfied.

Assume now that, for some ordinal $\alpha$, we have shown that $A$ satisfies
(WF) with respect to $M$ and $l$ for all $n\leq\omega$ and all $A\in B_P$
with $l(A)\leq (\alpha,n)$.

Let $A\in L_{\alpha+1}\setminus
L_{\alpha}=T_{P/G_{\alpha}}\upa\omega\setminus L_{\alpha}$. Then $A\in
T_{P/G_{\alpha}}\upa n\setminus L_{\alpha}$ for some $n\in\Nat$; note that
all (negative) literals which were removed by the Gelfond-Lifschitz
transformation from clauses with head $A$ have level less than
$(\alpha,0)$. Then the assertion that $A$ satisfies (WF) with respect to $M$
and $l$ follows again by Theorem \ref{theo:defleast}.

Let $A\in (B_P\setminus G_{\alpha+1})\cap G_{\alpha}$. Then $A\not\in
T_{P/L_{\alpha}}\upa\omega$. Now for any clause $A\gets A_1,\dots,A_k,\lnot
B_1,\dots,\lnot B_m$ in $\ground(P)$, if $B_j\in L_{\alpha}$ for some $j$, then
$l(A)>l(B_j)$. Otherwise, since $A\not\in T_{P/L_{\alpha}}\upa\omega$, we
have that there exists $A_i$ with $A_i\not\in T_{P/L_{\alpha}}\upa\omega$,
and hence $l(A)\geq l(A_i)$, and this suffices.

This finishes the proof that $P$ satisfies (WF) with respect to $M$ and
$l$. It therefore only remains to show that $M$ is greatest with this
property.

So assume that $M_1\not=M$ is the greatest model such that $P$ satisfies
(WF) with respect to $M_1$ and some $M_1$-partial level mapping $l_1$. 

Assume $L\in M_1\setminus M$ and, without loss of generality, let the
literal $L$ be chosen such that $l_1(L)$ is minimal. We consider the
following two cases.

(Case i) If $L=A$ is an atom, then there exists a clause $A\gets\body$ in
$\ground(P)$ such that $l_1(L)<l_1(A)$ for all literals $L$ in $\body$, and
such that $\body$ is true in $M_1$. Hence, $\body$ is true in $M$ and
$A\gets\body$ transforms to a clause $A\gets A_1,\dots, A_n$ in $P/G_P$ with
$A_1,\dots,A_n\in L_P=T_{P/G_P}\upa\omega$. But this implies $A\in M$,
contradicting $A\in M_1\setminus M$.

(Case ii) If $L=\lnot A\in M_1\setminus M$ is a negated atom, then $\lnot
A\in M_1$ and $A\in G_P=T_{P/L_P}\upa\omega$, so $A\in T_{P/L_P}\upa n$ for
some $n\in\Nat$. We show by induction on $n$ that this leads to a
contradiction, to finish the proof.

If $A\in T_{P/L_P}\upa 1$, then there is a unit clause $A\gets$ in $P/L_P$,
and any corresponding clause $A\gets\lnot B_1,\dots,\lnot B_k$ in
$\ground(P)$ satisfies $B_1,\dots,B_k\not\in L_P$. Since $\lnot A\in M_1$,
we also obtain by Theorem \ref{theo:wfchar} that there is
$i\in\{1,\dots,k\}$ such that $B_i\in M_1$ and $l_1(B_i)<l_1(A)$. By
minimality of $l_1(A)$, we obtain $B_i\in M$, and hence $B_i\in L_P$, which
contradicts $B_i\not\in L_P$.

Now assume that there is no $\lnot B\in M_1\setminus M$ with $B\in
T_{P/L_P}\upa k$ for any $k<n+1$, and let $\lnot A\in M_1\setminus M$ with
$A\in T_{P/L_P}\upa (n+1)$. Then there is a clause $A\gets A_1,\dots, A_m$
in $P/L_P$ with $A_1,\dots, A_m\in T_{P/L_P}\upa n\subseteq G_P$, and we
note that we cannot have $\lnot A_i\in M_1\setminus M$ for any
$i\in\{1,\dots,m\}$, by our current induction hypothesis. Furthermore, it is
also impossible for $\lnot A_i$ to belong to $M$ for any $i$, otherwise we
would have $A_i\in B_P\setminus G_P$. Thus, we conclude that we cannot have
$\lnot A_i\in M_1$ for any $i$. Moreover, there is a corresponding clause
$A\gets A_1,\dots,A_m,\lnot B_1,\dots,\lnot B_{m_1}$ in $\ground(P)$ with
$B_1,\dots, B_{m_1}\not\in L_P$. Hence, by Theorem \ref{theo:wfchar}, we
know that there is $i\in\{1,\dots, m_1\}$ such that $B_i\in M_1$ and
$l_1(B_i)<l_1(A)$. By minimality of $l_1(A)$, we conclude that $B_i\in M$,
so that $B_i\in L_P$, and this contradicts $B_i\not\in L_P$.
\end{proof}

\begin{bsp}\mylabel{bsp:afp2}
Consider again the program $P$ from Examples \ref{bsp:wf}, \ref{bsp:afp},
and \ref{bsp:wf2}. With notation from the proof of Theorem \ref{theo:altfp}
we get $l(q)=(1,0)$, $l(s)=(1,1)$, and $l(p)=(0,\omega)$. 
\end{bsp}

\section{Weakly Perfect Model Semantics}\mylabel{sec:wstrat}

By applying our proof scheme, we have obtained new and uniform
characterizations of the Fitting semantics and the well-founded semantics,
and argued that the well-founded semantics is a stratified version of the
Fitting semantics. Our argumentation is based on the key intuition
underlying the notion of stratification, that recursion should be allowed
through positive dependencies, but be forbidden through negative
dependencies. As we have seen in Theorem \ref{theo:wfchar}, the well-founded
semantics provides this for a setting in three-valued logic. Historically, a
different semantics, given by the so-called weakly perfect model associated
with each program, was proposed by Przymusinska and Przymusinski \cite{PP90}
in order to carry over the intuition underlying the notion of stratification
to a three-valued setting. In the following, we will characterize weakly
perfect models via level mappings, in the spirit of our approach. We will
thus have obtained uniform characterizations of the Fitting semantics, the
well-founded semantics, and the weakly perfect model semantics, which makes
it possible to easily compare them.

\begin{defn}\mylabel{def:ws}
Let $P$ be a normal logic program, I be a model of $P$ and $l$ be an
$I$-partial level mapping for $P$. We say that $P$ \emph{satisfies} (WS)
\emph{with respect to $I$ and $l$}, if each $A\in\domain(l)$ satisfies one of
the following conditions.
\begin{enumerate}
\item[(WSi)] $A\in I$ and there exists a clause $A\gets L_1,\dots,
L_n\in\ground(P)$ such that $L_i\in I$ and $l(A)>l(L_i)$ for all
$i=1,\dots,n$.
\item[(WSii)] $\lnot A\in I$ and for each clause $A\gets A_1,\dots,A_n,\lnot
B_1,\dots,\lnot B_m\in\ground(P)$ (at least) one of the following three
conditions holds.
\begin{enumerate}[({WSii}a)]
\item[(WSiia)] There exists $i$ such that $\lnot A_i\in I$ and $l(A)>l(A_i)$.
\item[(WSiib)] For all $k$ we have $l(A)\geq l(A_k)$, for all $j$ we have
$l(A)>l(B_j)$, and there exists $i$ with $\lnot A_i\in I$.
\item[(WSiic)] There exists $j$ such that $B_j\in I$ and $l(A)>l(B_j)$.
\end{enumerate}
\end{enumerate}
\end{defn}

We observe that the condition (WSii) in the above theorem is more general
than (Fii), and more restrictive than (WFii). 

We will see below in Theorem \ref{theo:wschar}, that Definition \ref{def:ws}
captures the weakly perfect model, in the same way in which Definitions
\ref{def:fit} and \ref{def:wfchar} capture the Fitting model, respectively
the well-founded model.

In order to proceed with this, we first need to recall the definition of
weakly perfect models due to Przymusinska and Przymusinski \cite{PP90}, and
we will do this next. For ease of notation, it will be convenient to
consider (countably infinite) propositional programs instead of programs
over a first-order language. This is both common practice and no
restriction, because the ground instantiation $\ground(P)$ of a given
program $P$ can be understood as a propositional program which may consist
of a countably infinite number of clauses. Let us remark that our definition
below differs slightly from the original one, and we will return to this
point later. It nevertheless leads to exactly the same notion of weakly
stratified program.

Let $P$ be a (countably infinite propositional) normal logic program. An
atom $A\in B_P$ \emph{refers to} an atom $B\in B_P$ if $B$ or $\lnot B$
occurs as a body literal in a clause $A\gets\body$ in $P$. $A$ \emph{refers
negatively to} $B$ if $\lnot B$ occurs as a body literal in such a
clause. We say that $A$ \emph{depends on} $B$ if the pair $(A,B)$ is in the
transitive closure of the relation \emph{refers to}, and we write this as
$B\leq A$. We say that $A$ \emph{depends negatively on} $B$ if there are
$C,D\in B_P$ such that $C$ refers negatively to $D$ and the following hold:
(1) $C\leq A$ or $C=A$ (the latter meaning identity). (2) $B\leq D$ or
$B=D$. We write $B<A$ in this case. For $A,B\in B_P$, we write $A\sim B$ if
either $A=B$, or $A$ and $B$ depend negatively on each other, i.e. if $A<B$
and $B<A$ both hold. The relation $\sim$ is an equivalence relation and its
equivalence classes are called \emph{components} of $P$. A component is
\emph{trivial} if it consists of a single element $A$ with $A\not< A$.

Let $C_1$ and $C_2$ be two components of a program $P$. We write $C_1\prec
C_2$ if and only if $C_1\not= C_2$ and for all $A_1\in C_1$ there is $A_2\in
C_2$ with $A_1<A_2$. A component $C_1$ is called \emph{minimal} if there is
no component $C_2$ with $C_2\prec C_1$.

Given a normal logic program $P$, the \emph{bottom stratum} $S(P)$ of $P$ is
the union of all minimal components of $P$. The \emph{bottom layer} of $P$
is the subprogram $L(P)$ of $P$ which consists of all clauses from $P$ with
heads belonging to $S(P)$. 

Given a (partial) interpretation $I$ of $P$, we define the \emph{reduct of
$P$ with respect to $I$} as the program $P/I$ obtained from $P$ by
performing the following reductions. (1) Remove from $P$ all clauses
which contain a body literal $L$ such that $\lnot L\in I$ or whose head
belongs to $I$. (2) Remove from all remaining clauses all body literals $L$
with $L\in I$. (3) Remove from the resulting program all non-unit clauses,
whose heads appear also as unit clauses in the program.

\begin{defn}\mylabel{def:wpm}
The \emph{weakly perfect model} $M_P$ of a program $P$ is defined by
transfinite induction as follows. Let $P_0=P$ and $M_0=\emptyset$. For each
(countable) ordinal $\alpha>0$ such that programs $P_\delta$ and partial
interpretations $M_\delta$ have already been defined for all
$\delta<\alpha$, let
\begin{displaymath}
\begin{array}{lcl}
N_\alpha &= &\bigcup_{0<\delta<\alpha} M_\delta,\\
P_\alpha &= &P/N_\alpha,\\
R_\alpha &\text{ is}& \text{ the set of all atoms which are undefined in }
N_\alpha\\
 &\phantom{=} &\text{and were eliminated from } P \text{ by reducing it with
respect to } N_\alpha,\\
S_\alpha &=&S\left(P_\alpha\right), \text{ and}\\
L_\alpha &=& L\left(P_\alpha\right).
\end{array}
\end{displaymath}

The construction then proceeds with one of the following three cases. (1) If
$P_\alpha$ is empty, then the construction stops and $M_P=N_\alpha\cup \lnot
R_\alpha$ is the (\emph{total}) \emph{weakly perfect model} of $P$. (2) If
the bottom stratum $S_\alpha$ is empty or if the bottom layer $L_\alpha$
contains a negative literal, then the construction also stops and $M_P=
N_\alpha\cup\lnot R_\alpha$ is the (\emph{partial}) \emph{weakly perfect
model} of $P$.  (3) In the remaining case $L_\alpha$ is a definite program,
and we define $M_\alpha=H\cup\lnot R_\alpha$, where $H$ is
the definite (partial) model of $L_\alpha$, and the construction continues.

For every $\alpha$, the set $S_\alpha\cup R_\alpha$ is called the
\emph{$\alpha$-th stratum} of $P$ and the program $L_\alpha$ is called the
\emph{$\alpha$-th layer} of $P$.
\end{defn}

A \emph{weakly stratified program} is a program with a total weakly perfect
model. The set of its strata is then called its \emph{weak stratification}. 

\begin{bsp}\mylabel{bsp:ws}
Consider the program $P$ which consists of the following six clauses.

\begin{center}
\begin{minipage}[t]{4cm}
\begin{displaymath}
\begin{array}{lcl}
a &\gets&\lnot b\\
b &\gets &c,\lnot a\\
b &\gets &c,\lnot d\\
c &\gets &b,\lnot e\\
d &\gets &e\\
e &\gets &d
\end{array}
\end{displaymath}
\end{minipage}
\end{center}

Then $N_1=M_1=\{\lnot d,\lnot e\}$ and $P/N_1$ consists of the clauses

\begin{center}
\begin{minipage}[t]{3.5cm}
\begin{displaymath}
\begin{array}{lcl}
a &\gets&\lnot b\\
b &\gets &c,\lnot a\\
b &\gets &c\\
c &\gets &b .
\end{array}
\end{displaymath}
\end{minipage}
\end{center}

Its least component is $\{a,b,c\}$. The corresponding bottom layer, which is
all of $P/N_1$, contains a negative literal, so the construction stops and
$M_2=N_1=\{\lnot d,\lnot e\}$ is the (partial) weakly perfect model of $P$.
\end{bsp}

Let us return to the remark made earlier that our definition of weakly
perfect model, as given in Definition \ref{def:wpm}, differs slightly from
the version introduced by Przymusinska and Przymusinski \cite{PP90}. In
order to obtain the original definition, points (2) and (3) of Definition
\ref{def:wpm} have to be replaced as follows: (2) If the bottom stratum
$S_\alpha$ is empty or if the bottom layer $L_\alpha$ \emph{has no least
two-valued model}, then the construction stops and $M_P= N_\alpha\cup\lnot
R_\alpha$ is the (partial) weakly perfect model of $P$. (3) In the remaining
case $L_\alpha$ \emph{has a least two-valued model}, and we define
$M_\alpha=H\cup\lnot R_\alpha$, where $H$ is the partial model of $L_\alpha$
corresponding to its least two-valued model, and the construction continues.

The original definition is more general due to the fact that every definite
program has a least two-valued model. However, while the least two-valued
model of a definite program can be obtained as the least fixed point of the
monotonic (and even Scott-continuous) operator $T_P^\plus$, we know of no
similar result, or general operator, for obtaining the least two-valued
model, if existent, of progams which are not definite. The original
definition therefore seems to be rather awkward, and indeed, for the
definition of weakly stratified programs \cite{PP90}, the more general
version was dropped in favour of requiring definite layers. So Definition
\ref{def:wpm} is an adaptation taking the original notion of weakly
stratified program into account, and appears to be more natural. In the
following, the notion of \emph{weakly perfect model} will refer to
Definition \ref{def:wpm}.

To be pedantic, there is another difference, namely that we have made
explicit the sets $R_\alpha$ of Definition \ref{def:wpm}, which were only
implicitly treated in the original definition. The result is the same.

We show next that Definition \ref{def:ws} indeed captures the weakly
perfect model. The proof basically follows our proof scheme, with some
alterations, and the monotonic construction which defines the weakly perfect
model serves in place of a monotonic operator. The technical details of the
proof are very involved.

\begin{theo}\mylabel{theo:wschar}
Let $P$ be a normal logic program with weakly perfect model $M_P$. Then
$M_P$ is the greatest model among all models $I$, for which there exists an
$I$-partial level mapping $l$ for $P$ such that $P$ satisfies (WS) with
respect to $I$ and $l$.
\end{theo}

We prepare the proof of Theorem \ref{theo:wschar} by introducing some
notation, which will make the presentation much more transparent. As for the
proof of Theorem \ref{theo:altfp}, we will consider level mappings which map
into pairs $(\beta,n)$ of ordinals, where $n\leq\omega$.

Let $P$ be a normal logic program with (partial) weakly perfect model
$M_P$. Then define the $M_P$-partial level mapping $l_P$ as follows:
$l_P(A)=(\beta,n)$, where $A\in S_\beta\cup R_\beta$ and $n$ is least with
$A\in T_{L_\beta}^\plus\upa (n+1)$, if such an $n$ exists, and $n=\omega$
otherwise. We observe that if $l_P(A)=l_P(B)$ then there exists $\alpha$
with $A,B\in S_\alpha\cup R_\alpha$, and if $A\in S_\alpha\cup R_\alpha$ and
$B\in S_\beta\cup R_\beta$ with $\alpha<\beta$, then $l(A)<l(B)$.

The following definition is again technical and will help to ease notation
and arguments.

\begin{defn}\mylabel{def:mcs}
 Let $P$ and $Q$ be two programs and let $I$ be an interpretation.
\begin{enumerate}
\item If $C_1=(A\gets L_1,\dots,L_m)$ and $C_2=(B\gets K_1,\dots,K_n)$ are
two clauses, then we say that \emph{$C_1$ subsumes $C_2$}, written
$C_1\preccurlyeq C_2$, if $A=B$ and
$\{L_1,\dots,L_m\}\subseteq\{K_1,\dots,K_n\}$.
\item We say that \emph{$P$ subsumes $Q$}, written $P\preccurlyeq Q$, if for
each clause $C_1$ in $P$ there exists a clause $C_2$ in $Q$ with
$C_1\preccurlyeq C_2$.
\item We say that \emph{$P$ subsumes $Q$ model-consistently} (\emph{with
respect to $I$}), written $P\preccurlyeq_I Q$, if the following conditions
hold. (i) For each clause $C_1=(A\gets L_1,\dots,L_m)$ in $P$ there exists a
clause $C_2=(B\gets K_1,\dots,K_n)$ in $Q$ with $C_1\preccurlyeq C_2$ and
$(\{K_1,\dots,K_n\}\setminus\{L_1,\dots,L_m\})\subseteq I$. (ii) For each
clause $C_2=(B\gets K_1,\dots,K_n)$ in $Q$ with $\{K_1,\dots,K_n\}\in I$ and
$B\not\in I$ there exists a clause $C_1$ in $P$ such that $C_1\preccurlyeq
C_2$.
\end{enumerate}
\end{defn}

A clause $C_1$ subsumes a clause $C_2$ if both have the same head and the
body of $C_2$ contains at least the body literals of $C_1$, e.g. $p\gets q$
subsumes $p\gets q,\lnot r$. A program $P$ subsumes a program $Q$ if every
clause in $P$ can be generated this way from a clause in $Q$, e.g. the
program consisting of the two clauses $p\gets q$ and $p\gets r$ subsumes the
program consisting of $p\gets q,\lnot s$ and $p\gets r,p$. This is also an
example of a model-consistent subsumption with respect to the interpretation
$\{\lnot s,p\}$. Concerning Example \ref{bsp:ws}, note that
$P/N_1\preccurlyeq_{N_1} P$, which is no coincidence. Indeed, Definition
\ref{def:mcs} facilitates the proof of Theorem \ref{theo:wschar} by
employing the following lemma.

\begin{lemma}\mylabel{lem:mcs}
With notation from Definiton \ref{def:wpm}, we have
$P/N_\alpha\preccurlyeq_{N_\alpha} P$ for all $\alpha$. 
\end{lemma}

\begin{proof}
Condition 3(i) of Definition \ref{def:mcs} holds because every clause in
$P/N_\alpha$ is obtained from a clause in $P$ by deleting body literals
which are contained in $N_\alpha$. Condition 3(ii) holds because for each
clause in $P$ with head $A\not\in N_\alpha$ whose body is true under
$N_\alpha$, we have that $A\gets$ is a fact in $P/N_\alpha$.
\end{proof}

The next lemma establishes the induction step in part (2) of the proof of
Theorem \ref{theo:wschar}.

\begin{lemma}\mylabel{lem:main}
If $I$ is a non-empty model of a (infinite propositional normal) logic
program $P'$ and $l$ an $I$-partial level mapping such that $P'$ satisfies
(WS) with respect to $I$ and $l$, then the following hold for
$P=P'/\emptyset$.
\begin{enumerate}[(a)]
\item The bottom stratum $S(P)$ of $P$ is non-empty and consists of trivial
components only.
\item The bottom layer $L(P)$ of $P$ is definite.
\item The definite (partial) model $N$ of $L(P)$ is consistent with $I$ in
the following sense: we have $I'\subseteq N$, where $I'$ is the restriction
of $I$ to all atoms which are not undefined in $N$.
\item $P/N$ satisfies (WS) with respect to $I\setminus N$ and $l/N$, where
$l/N$ is the restriction of $l$ to the atoms in $I\setminus N$.
\end{enumerate}
\end{lemma}

\begin{proof}
(a) Assume there exists some component $C\subseteq S(P)$ which is not
trivial. Then there must exist atoms $A,B\in C$ with $A<B$, $B<A$, and
$A\not=B$. Without loss of generality, we can assume that $A$ is chosen such
that $l(A)$ is minimal. Now let $A'$ be any atom occuring in a clause with
head $A$. Then $A>B>A\geq A'$, hence $A>A'$, and by minimality of the
component we must also have $A'>A$, and we obtain that all atoms occuring in
clauses with head $A$ must be contained in $C$. We consider two cases.

(Case i) If $A\in I$, then there must be a fact $A\gets$ in $P$, since
otherwise by (WSi) we had a clause $A\gets L_1,\dots,L_n$ (for some $n\geq
1$) with $L_1,\dots,L_n\in I$ and $l(A)>l(L_i)$ for all $i$, contradicting
the minimality of $l(A)$. Since $P=P'/\emptyset$ we obtain that $A\gets$ is
the only clause in $P$ with head $A$, contradicting the existence of
$B\not=A$ with $B<A$.

(Case ii) If $\lnot A\in I$, and since $A$ was chosen minimal with respect
to $l$, we obtain that condition (WSiib) must hold for each clause $A\gets
A_1,\dots,A_n,\lnot B_1,\dots,\lnot B_m$ with respect to $I$ and $l$, and
that $m=0$. Furthermore, all $A_i$ must be contained in $C$, as already
noted above, and $l(A)\geq l(A_i)$ for all $i$ by (WSiib). Also from (Case
i) we obtain that no $A_i$ can be contained in $I$. We have now established
that for all $A_i$ in the body of any clause with head $A$, we have
$l(A)=l(A_i)$ and $\lnot A_i\in I$. The same argument holds for all clauses
with head $A_i$, for all $i$, and the argument repeats. Now from $A>B$ we
obtain that there are $D,E\in C$ with $A\geq E$ (or $A=E$), $D\geq B$ (or
$D=B$), and $E$ refers negatively to $D$. As we have just seen, we obtain
$\lnot E\in I$ and $l(E)=l(A)$. Since $E$ refers negatively to $D$, there is
a clause with head $E$ and $\lnot D$ contained in the body of this
clause. Since (WSii) holds for this clause, there must be a literal $L$ in
the body with level less than $l(E)$, hence $l(L)<l(A)$ and $L\in C$ which
is a contradiction. We thus have established that all components are
trivial.

We show next that the bottom stratum is non-empty. Indeed, let $A$ be an
atom such that $l(A)$ is minimal. We will show that $\{A\}$ is a
component. So assume it is not, i.e. that there is $B$ with $B<A$. Then
there exist $D_1,\dots, D_k$, for some $k\in\Nat$, such that $D_1=A$, $D_j$
refers to $D_{j+1}$ for all $j=1,\dots,k-1$, and $D_k$ refers negatively to
some $B'$ with $B'\geq B$ (or $B'=B$).

We show next by induction that for all $j=1,\dots, k$ the following
statements hold: $\lnot D_j\in I$, $B<D_j$, and $l(D_j)=l(A)$. Indeed note
that for $j=1$, i.e. $D_j=A$, we have that $B<D_j=A$ and $l(D_j)=l(A)$.
Assuming $A\in I$, we obtain by minimality of $l(A)$ that $A\gets$ is the
only clause in $P=P'/\emptyset$ with head $A$, contradicting the existence
of $B<A$. So $\lnot A\in I$, and the assertion holds for $j=1$. Now assume
the assertion holds some $j<k$. Then obviously $D_{j+1}>B$. By $\lnot D_j\in
I$ and $l(D_j)=l(A)$, we obtain that (WSii) must hold, and by the minimality
of $l(A)$ we infer that (WSiib) must hold and that no clause with head $D_j$
contains negated atoms. So $l(D_{j+1})=l(D_j)=l(A)$ holds by (WSiib) and
minimality of $l(A)$. Furthermore, the assumption $D_{j+1}\in I$ can be
rejected by the same argument as for $A$ above, because then $D_{j+1}\gets$
would be the only clause with head $D_{j+1}$, by minimality of
$l(D_{j+1})=l(A)$, contradicting $B<D_{j+1}$. This concludes the inductive
proof.

Summarizing, we obtain that $D_k$ refers negatively to $B'$, and that $\lnot
D_k\in I$. But then there is a clause with head $D_k$ and $\lnot B'$ in its
body which satisfies (WSii), contradicting the minimality of
$l(D_k)=l(A)$. This concludes the proof of statement (a).

(b) According to \cite{PP90} we have that whenever all components are
trivial, then the bottom layer is definite. So the assertion follows from
(a).

(c) Let $A\in I'$ be an atom with $A\not\in N$, and assume without loss of
generality that $A$ is chosen such that $l(A)$ is minimal with these
properties. Then there must be a clause $A\gets\body$ in $P$ such that all
literals in $\body$ are true with respect to $I'$, hence with respect to $N$
by minimality of $l(A)$. Thus $\body$ is true in $N$, and since $N$ is a
model of $L(P)$ we obtain $A\in N$, which contradicts our assumption.

Now let $A\in N$ be an atom with $A\not\in I'$, and assume without loss of
generality that $A$ is chosen such that $n$ is minimal with $A\in
T_{L(P)}^\plus\upa (n+1)$. But then there is a definite clause $A\gets\body$ in
$L(P)$ such that all atoms in $\body$ are true with respect to
$T_{L(P)}^\plus\upa n$, hence also with respect to $I'$, and since $I'$ is a
model of $L(P)$ we obtain $A\in I'$, which contradicts our
assumption. 

Finally, let $\lnot A\in I'$. Then we cannot have $A\in N$ since this
implies $A\in I'$. So $\lnot A\in N$ since $N$ is a total model of $L(P)$.

(d) From Lemma \ref{lem:mcs}, we know that $P/N\preccurlyeq_N P$. We
distinguish two cases. 

(Case i) If $A\in I\setminus N$, then there must exist a clause
$A\gets L_1,\dots, L_k$ in $P$ such that $L_i\in I$ and $l(A)>l(L_i)$ for
all $i$. Since it is not possible that $A\in N$, there must also be a clause
in $P/N$ which subsumes $A\gets L_1,\dots,L_k$, and which therefore
satisfies (WSi). So $A$ satisfies (WSi).

(Case ii) If $\lnot A\in I\setminus N$, then for each clause
$A\gets\bodyone$ in $P/N$ there must be a clause $A\gets\body$ in $P$ which
is subsumed by the former, and since $\lnot A\in I$, we obtain that
condition (WSii) must be satisfied by $A$, and by the clause
$A\gets\body$. Since reduction with respect to $N$ removes only body
literals which are true in $N$, condition (WSii) is still met.
\end{proof}

We can now proceed with the proof.

\bigskip
\begin{proof}[Proof of Theorem \ref{theo:wschar}]
The proof will be established by showing the following facts: (1) $P$
satisfies (WS) with respect to $M_P$ and $l_P$. (2) If $I$ is a model of $P$ 
and $l$ an $I$-partial level mapping such that $P$ satisfies (WS) with
respect to $I$ and $l$, then $I\subseteq M_P$.

(1) Let $A\in\domain(l_P)$ and $l_P(A)=(\alpha,n)$. We consider two cases.

(Case i) If $A\in M_P$, then $A\in T_{L_\alpha}^\plus\upa (n+1)$. Hence there
exists a definite clause $A\gets A_1,\dots,A_k$ in $L_\alpha$ with
$A_1,\dots,A_k\in T_{L_\alpha}^\plus\upa n$, so $A_1,\dots, A_k\in M_P$ with
$l_P(A)>l_P(A_i)$ for all $i$. Since $P/N_\alpha\preccurlyeq_{N_\alpha} P$
by Lemma \ref{lem:mcs}, there must exist a clause $A\gets A_1,\dots,
A_k,L_1,\dots,L_m$ in $P$ with literals $L_1,\dots,L_m\in N_\alpha\subseteq
M_P$, and we obtain $l_P(L_j)<l_P(A)$ for all $j=1,\dots,m$. So (WSi) holds
in this case.

(Case ii) If $\lnot A\in M_P$, then let $A\gets A_1,\dots, A_k,\lnot
B_1,\dots,\lnot B_m$ be a clause in $P$, noting that (WSii) is trivially
satisfied in case no such clause exists. We consider the following two
subcases.

(Subcase ii.a) Assume $A$ is undefined in $N_\alpha$ and was eliminated from
$P$ by reducing it with respect to $N_\alpha$, i.e. $A\in R_\alpha$. Then,
in particular, there must be some $\lnot A_i\in N_\alpha$ or some $B_j\in
N_\alpha$, which yields $l_P(A_i)<l_P(A)$, respectively $l_P(B_j)<l_P(A)$,
and hence one of (WSiia), (WSiic) holds.

(Subcase ii.b) Assume $\lnot A\in H$, where $H$ is the definite (partial)
model of $L_\alpha$. Since $P/N_\alpha$ subsumes $P$ model-consistently with
respect to $N_\alpha$, we obtain that there must be some $A_i$ with $\lnot
A_i\in H$, and by definition of $l_P$ we obtain
$l_P(A)=l_P(A_i)=(\alpha,\omega)$, and hence also $l_P(A_{i'})\leq l_P(A_i)$ for
all $i'\not=i$. Furthermore, since $P/N_\alpha$ is definite, we obtain that
$\lnot B_j\in N_\alpha$ for all $j$, hence $l_P(B_j)<l_P(A)$ for all $j$. So
condition (WSiib) is satisfied.

(2) First note that for all models $M$, $N$ of $P$ with $M\subseteq N$ we
have $(P/M)/N=P/(M\cup N)= P/N$ and $(P/N)/\emptyset = P/N$. 

Let $I_\alpha$ denote $I$ restricted to the atoms which are not undefined in
$N_\alpha\cup R_\alpha$. It suffices to show the following: For all
$\alpha>0$ we have $I_\alpha\subseteq N_\alpha\cup R_\alpha$, and
$I\setminus M_P=\emptyset$.

We next show by induction that if $\alpha>0$ is an ordinal, then the
following statements hold. (a) The bottom stratum of $P/N_\alpha$ is
non-empty and consists of trivial components only. (b) The bottom layer of
$P/N_\alpha$ is definite. (c) $I_\alpha\subseteq N_\alpha\cup R_\alpha$. (d)
$P/N_{\alpha+1}$ satisfies (WS) with respect to $I\setminus N_{\alpha+1}$
and $l/N_{\alpha+1}$.

Note first that $P$ satisfies the hypothesis of Lemma \ref{lem:main}, hence
also its consequences. So $P/N_1=P/\emptyset$ satisfies (WS) with respect to
$I\setminus N_1$ and $l/N_1$, and by application of Lemma \ref{lem:main} we
obtain that statements (a) and (b) hold. For (c), note that no atom in $R_1$
can be true in $I$, because no atom in $R_1$ can appear as head of a clause
in $P$, and apply Lemma \ref{lem:main} (c). For (d), apply Lemma
\ref{lem:main}, noting that $P/N_2\preccurlyeq_{N_2} P$.

For $\alpha$ being a limit ordinal, we can show exactly as in the proof of
Lemma \ref{lem:main} (d), that $P$ satisfies (WS) with respect to
$I\setminus N_{\alpha}$ and $l/N_\alpha$. So Lemma \ref{lem:main} is
applicable and statements (a) and (b) follow. For (c), let $A\in
R_\alpha$. Then every clause in $P$ with head $A$ contains a body literal
which is false in $N_\alpha$. By induction hypothesis, this implies that no
clause with head $A$ in $P$ can have a body which is true in $I$. So
$A\not\in I$. Together with Lemma \ref{lem:main} (c), this proves statement
(c). For (d), apply again Lemma \ref{lem:main} (d), noting that
$P/N_{\alpha+1}\preccurlyeq_{N_{\alpha+1}} P$.

For $\alpha=\beta+1$ being a successor ordinal, we obtain by induction
hypothesis that $P/N_\beta$ satisfies the hypothesis of Lemma
\ref{lem:main}, so again statements (a) and (b) follow immediately from this
lemma, and (c), (d) follow as in the case for $\alpha$ being a limit
ordinal.

It remains to show that $I\setminus M_P=\emptyset$. Indeed by the
transfinite induction argument just given we obtain that $P/M_P$ satisfies
(WS) with respect to $I\setminus M_P$ and $l/M_P$. If $I\setminus M_P$ is
non-empty, then by Lemma \ref{lem:main} the bottom stratum $S(P/M_P)$ is
non-empty and the bottom layer $L(P/M_P)$ is definite with definite
(partial) model $M$. Hence by definition of the weakly perfect model $M_P$
of $P$ we must have that $M\subseteq M_P$ which contradicts the fact that
$M$ is the definite model of $L(P/M_P)$. Hence $I\setminus M_P$ must be
empty which concludes the proof.
\end{proof}

Of independent interest is again the case, where the model in question is
total. We see immediately, for example, in the light of Theorem
\ref{theo:stablechar}, that the model is then stable.

\begin{kor}
A normal logic program $P$ is weakly stratified, i.e. has a total weakly
perfect model, if and only if there is a total model $I$ of $P$ and a
(total) level mapping $l$ for $P$ such that $P$ satisfies (WS) with respect
to $I$ and $l$.
\end{kor}

We also obtain the following corollary as a trivial consequence of our
uniform characterizations by level mappings.

\begin{kor}\mylabel{kor:compfitwswf}
Let $P$ be a normal logic progam with Fitting model $M_{\text{F}}$, weakly
perfect model $M_{\text{WP}}$, and well-founded model $M_{\text{WF}}$. Then
$M_{\text{F}}\subseteq M_{\text{WP}}\subseteq M_{\text{WF}}$.
\end{kor}

\begin{bsp}\mylabel{bsp:ws2}
Consider the program $P$ from Example \ref{bsp:ws}. Then
$M_{\text{F}}=\emptyset$, $M_{\text{WP}}=\{\lnot d,\lnot e\}$, and
$M_{\text{WF}}=\{a,\lnot b,\lnot c,\lnot d,\lnot e\}$.
\end{bsp}

\section{Related Work}\mylabel{sec:related}

As already mentioned in the introduction, level mappings have been used for
studying semantic aspects of logic programs in a number of different
ways. Our presentation suggests a novel application of level mappings,
namely for providing uniform characterizations of different fixed-point
semantics for logic programs with negation. Although we believe our
perspective to be new in this general form, there nevertheless have been
results in the literature which are very close in spirit to our
characterizations. 

A first noteable example of this is Fages' characterization of stable models
\cite{Fag94}, which we have stated in Theorem \ref{theo:stablechar}.
Another result which uses level mappings to characterize a semantics is by
Lifschitz, Przymusinski, St\"ark, and McCain \cite[Lemma 3]{LCPS95}. We
briefly compare their characterization of the well-founded semantics and
ours. In fact, this discussion can be based upon two different
characterizations of the least fixed point of a monotonic operator $F$. On
the one hand, this least fixed point is of course the \emph{least of all
fixed points} of $F$, and on the other hand, this least fixed point is the
limit of the sequence of powers $(F\uparrow \alpha)_\alpha$, and in this
latter sense is the \emph{least iterate} of $F$ \emph{which is also a fixed
point}. Our characterizations of definite, Fitting, well-founded, and weakly
stratified semantics use the latter approach, which is reflected in our
general proof scheme, which defines level mappings according to powers, or
iterates, of the respective operators. The results by Fages \cite{Fit94} and
Lifschitz et al. \cite{LCPS95} hinge upon the former approach, i.e. they are
based on the idea of characterizing the fixed points of an operator ---
$\GL_P$, respectively $\Psi_P$ \cite{Prz89,BNN91} --- and so the sought
fixed point turns out to be the least of those. Consequently, as can be seen
in the proof of Theorem \ref{theo:stablechar}, the level mapping in Fages'
characterization, and likewise in the result by Lifschitz et al., arises
only indirectly from the operator --- $\GL_P$, respectively $\Psi_P$ ---
whose fixed point is sought. Indeed, the level mapping by Fages is defined
according to iterates of $T_{P/I}$, which is the operator for obtaining
$\GL_P(I)$, for any $I$. The result by Lifschitz et al. is obtained
similarly based on a three-valued operator $\Psi_P$.

Unforunately, these characterizations by Fages, in Theorem
\ref{theo:stablechar}, respectively by Lifschitz et al. \cite{LCPS95}, seem
to be applicable only to operators which are defined by least fixed points
of other operators, as is the case for $\GL_P$ and $\Psi_P$, and it seems
that the approach by Lifschitz et al. is unlikely to scale to other
semantics. For example, we attempted a straightforward characterization of
the Fitting semantics in the spririt of Lifschitz et al. which failed.

On a more technical level, a difference between our result, Theorem
\ref{theo:wfchar}, and the characterization by Lifschitz et
al. \cite{LCPS95} of the well-founded semantics is this: In our
characterization, the model is described using conditions on atoms which are
true or false (i.e. \emph{not undefined}) in the well-founded model, whereas
in theirs the conditions are on those atoms which are true or undefined
(i.e. \emph{not false}) in the well-founded model. The reason for this is
that we consider iterates of $W_P$, where $W_P\uparrow 0=\emptyset$, while
they use the fact that each fixed point of $\Psi_P$ is a least fixed point
of $\Phi_{P/I}$ with respect to the truth ordering on interpretations (note
that in this case $P/I$ denotes a three-valued generalization of the
Gelfond-Lifschitz transformation due to Przymusinski \cite{Prz89}). In this
ordering we have $\Phi_{P/I}\uparrow 0=\neg B_P$. It is nevertheless nice to
note that in the special case of the well-founded semantics there exist two
complementary characterizations using level mappings.

Since our proposal emphasizes uniformity of characterizations, it is related
to the large body of work on uniform approaches to logic programming
semantics, of which we will discuss two in more detail: the algebraic
approach via bilattices due to Fitting, and the work of Dix.

Bilattice-based semantics has a long tradition in logic programming theory,
starting out from the four-valued logic of Belnap \cite{Bel77}. The
underlying set of truth values, a four-element lattice, was recognized to
admit two ordering relations which can be interpreted as truth- and
knowledge-order. As such it has the structure of a bilattice, a term due to
Ginsberg \cite{Gin86}, who was the first to note the importance of
bilattices for inference in artificial intelligence \cite{Gin88}. This
general approach was imported into logic programming theory by Fitting
\cite{Fit91}.  Although multi-valued logics had been used for logic
programming semantics before \cite{Fit85}, bilattices provided an
interesting approach to semantics as they are capable of incorporating both
reasoning about truth and reasoning about knowledge, and, more technically,
because they have nice algebraic behaviour. Using this general framework
Fitting was able to show interesting relationships between the stable and
the well-founded semantics \cite{Fit91a,Fit93,FitTCS}.

Without claiming completeness we note two current developments in the
bilattice-based approach to logic programming: Fitting's framework has been
extended to an algebraic approach for approximating operators by Denecker,
Marek, and Truszczynski \cite{DMT00}. The inspiring starting point of this
work was the noted relationship between the stable model semantics and the
well-founded semantics, the latter approximating the former. The other line
of research was pursued mainly by Arieli and Avron \cite{AA94,AA98,Ari02},
who use bilattices for paraconsistent reasoning in logic programming. The
above outline of the historical development of bilattices in logic
programming theory suggests a similar kind of uniformity as we claim for our
approach. The exact relationship between both approaches, however, is still
to be investigated. On the one hand, bilattices can cope with
paraconsistency --- an issue of logic programming and deductive databases,
which is becoming more and more important --- in a very convenient way. On
the other hand, our approach can deal with semantics based on multi-valued
logics, whose underlying truth structure is not a bilattice. A starting
point for investigations in this direction could be the obvious meeting
point of both theories: the well-founded semantics for which we can provide
a characterization and which is a special case of the general approximation
theory of Denecker et al. \cite{DMT00}.

Another very general, and uniform, approach to logic programming pursues a
different point of view, namely logic programming semantics as nonmonotonic
inference. The general theory of nonmonotonic inference and a classification
of properties of nonmonotonic operators was developed by Kraus, Lehmann, and
Magidor \cite{KLM90}, leading to the notion \emph{KLM-axioms} for these
properties, and developed further by Makinson \cite{Mak94}. These axioms
were adopted to the terminology of logic programming and extended to a
general theory of logic programming semantics by Dix
\cite{Dix95a,Dix95b}. In this framework, different known semantics are
classified according to strong properties --- the KLM-axioms which hold for
the semantics -- and weak properties --- specific properties which deal with
the irregularities of negation-as-failure. As such Dix' framework is indeed
a general and uniform approach to logic programming, its main focus being on
semantic properties of logic programs. Our approach in turn could be called
\emph{semi-syntactic} in that definitions that employ level mappings
naturally take the structure of the logic program into account.  As in the
case of the bilattice-based approaches, it is not yet completely clear
whether these two approaches can be amalgamated in the sense of a
correspondence between properties of level mappings, e.g. strict or
semi-strict descent of the level, etc., on the one hand, and
KLM-properties of the logic program on the other. However, we believe that
it is possible to develop a proof scheme for nonmonotonic properties of
logic programs in the style of the proof scheme presented in the paper,
which can be used to cast semantics based on monotonic operators into level
mapping form.

We finally mention the work by Hitzler and Seda \cite{HS99a}, which was the
root and starting point for our investigations. This framework aims at the
characterization of program classes, such as (locally) stratified programs
\cite{ABW88,Prz88}, acceptable programs \cite{AP93}, or $\Phi$-accessible
programs \cite{HS99a}. Such program classes appear naturally whenever a
semantics is not defined for all logic programs. In these cases one tries to
characterize those programs, for which the semantics is well-defined or
well-behaved. Their main tool were monotonic operators in three-valued
logic, in the spirit of Fitting's $\Phi_P$, rather than level mappings. With
each operator comes a least fixed point, hence a semantics, and it is easily
checked that these semantics can be characterized using our approach, again
by straightforward application of our proof scheme. Indeed, preliminary
steps in this direction already led to an independent proof of a special
case of Corollary \ref{kor:compfitwswf} \cite{HS01b}.

\section{Conclusions and Further Work}\mylabel{sec:conc}

We have proposed a novel approach for obtaining uniform characterizations of
different semantics for logic programs. We have exemplified this by giving
new alternative characterizations of some of the major semantics from the
literature. We have developed and presented a methodology for obtaining
characterizations from monotonic semantic operators or related
constructions, and a proof scheme for showing correctness of the obtained
characterizations. We consider our contribution to be fundamental, with
potential for extension in many directions.

Our approach employs level mappings as central tool. The uniformity with
which our characterizations were obtained and proven to be correct suggests
that our method should be of wider applicability. In fact, since it builds
upon the well-known Tarski fixed point theorem, it should scale well to
most, if not all semantics, which are defined by means of a monotonic
operator. The main contribution of this paper is thus, that we have
developed a novel way of presenting logic programming semantics in some kind
of \emph{normal} or \emph{standard form}. This can be used for easy
comparison of semantics with respect to the syntactic structures that can be
used with a certain semantics, i.e. to what extent the semantics is able to
'break up' positive or negative dependencies or loops between atoms in the
program, as in Corollary \ref{kor:compfitwswf}.

However, there are many more requirements which a general and uniform
approach to logic program semantics should eventually be able to meet,
including (i) a better understanding of known semantics, (ii) proof schemes
for deriving properties of semantics, (iii) extendability to new programming
constructs, and (iv) support for designing new semantics for special
purposes.

Requirement (i) is met to some extent by our appoach, since it enables easy
comparison of semantics, as discussed earlier. However, in order to meet the
other requirements, i.e. to set up a \emph{meta-theory} of
level-mapping-based semantics, a lot of further research is needed. We list
some topics to be pursued in the future, some of which are under current
investigation by the authors. There are many properties which are
interesting to know about a certain semantics, depending on one's
perspective. For the nonmonotonic reasoning aspect of logic programming it
would certainly be interesting to have a proof scheme as flexible and
uniform as the one presented in this paper. Results and proofs in the
literature \cite{Fag94,Dix95a,Tur01} suggest that there is a strong
dependency between notions of ordering on the Herbrand base, as expressed by
level mappings, and KLM-properties satisfied by a semantics, which
constitutes some evidence that a general proof scheme for proving
KLM-properties from level mapping definitions can be developed. Other
interesting properties are e.g. the computational complexity of a semantics,
but also logical characterizations of the behaviour of negation in logic
programs, a line of research initiated by Pearce \cite{Pea97}.

For (iii), it would be desirable to extend our characterizations also to
disjunctive programs, which could perhaps contribute to the discussion about
appropriate generalizations of semantics of normal logic programs to the
disjunctive case.

We finally want to mention that the elegant mathematical framework of level
mapping definitions naturally gives rise to the design of new
semantics. However, at the time being this is only a partial fulfillment of
(iv): As long as a meta-theory for level-mapping-based semantics is missing,
one still has to apply conventional methods for extracting properties of the
respective semantics from its definition.

\bibliography{hitzler.bib}

\end{document}